\pdfoutput=1

\documentclass[11pt]{article}

\usepackage[]{EMNLP2023}

\usepackage{times}
\usepackage{latexsym}

\usepackage[T1]{fontenc}

\usepackage[utf8]{inputenc}

\usepackage{microtype}

\usepackage{inconsolata}

\usepackage{lipsum}
\usepackage{adjustbox}
\usepackage{graphicx}
\usepackage{pgf}

\usepackage{todonotes}
\usepackage{multirow, multicol}
\usepackage{amsmath, amssymb}
\usepackage{mathtools}
\DeclareMathOperator*{\argmax}{argmax}

\newcommand{\smid}{\!\mid\!}  
\newcommand{\BLEU}{\!\scalebox{.8}[1.0]{\textsc{Bleu}}\!}
\newcommand{\COMET}{\!\scalebox{.8}[1.0]{\textsc{Comet}}\!}
\newcommand{\nhphantom}[1]{\sbox0{#1}\hspace{-\the\wd0}}  
\usepackage{footnote}
\makesavenoteenv{tabular}
\makesavenoteenv{table}
\makesavenoteenv{table*}
\makeatletter
\newcommand\footnoteref[1]{\protected@xdef\@thefnmark{\ref{#1}}\@footnotemark}
\makeatother
\usepackage[normalem]{ulem}

\title{Document-Level Language Models for Machine Translation}

\author{Frithjof Petrick$^{1, 2}$ \quad Christian Herold$^{1, 2}$ \quad Pavel Petrushkov$^{1}$\\ \textbf{Shahram Khadivi$^{1}$ \quad Hermann Ney$^{2}$} \\
$^{1}$eBay, Inc., Aachen, Germany\\
\texttt{\{cherold, ppetrushkov, skhadivi\}@ebay.com} \\
$^{2}$Human Language Technology and Pattern Recognition Group \\
RWTH Aachen University, Aachen, Germany \\
\texttt{\{petrick, ney\}@i6.informatik.rwth-aachen.de} 
}

\begin{document}
\maketitle

\begin{abstract}
Despite the known limitations, most machine translation systems today still operate on the sentence-level.
One reason for this is, that most parallel training data is only sentence-level aligned, without document-level meta information available.
In this work, we set out to build context-aware translation systems utilizing document-level monolingual data instead.
This can be achieved by combining any existing sentence-level translation model with a document-level language model. 
We improve existing approaches by leveraging recent advancements in model combination.
Additionally, we propose novel weighting techniques that make the system combination more flexible and significantly reduce computational overhead.
In a comprehensive evaluation on four diverse translation tasks, we show that our extensions improve document-targeted scores substantially and are also computationally more efficient.
However, we also find that in most scenarios, back-translation gives even better results, at the cost of having to re-train the translation system.
Finally, we explore language model fusion in the light of recent advancements in large language models.
Our findings suggest that there might be strong potential in utilizing large language models via model combination.
\end{abstract}

\section{Introduction}

Machine translation (MT), the automatic translation of text from one language to another, has seen significant advancements in recent years, primarily driven by neural machine translation (NMT) models \cite{DBLP:journals/corr/BahdanauCB14, DBLP:conf/nips/VaswaniSPUJGKP17}.
These models have demonstrated remarkable capabilities in capturing complex linguistic patterns and producing high-quality translations \cite{DBLP:journals/corr/WuSCLNMKCGMKSJL16, DBLP:journals/corr/abs-1803-05567}.
Nevertheless, most models to-date operate on sentence-level, i.e. translate sentences independently without the context of the surrounding document.
Without access to such context, it is impossible for these MT systems to account for discourse-level phenomena such as resolution of ambiguous words and coherence.
Unsurprisingly, automatic translations are perceived as much worse, when they are evaluated on entire documents rather than just at the sentence-level \cite{DBLP:conf/emnlp/LaubliS018,DBLP:journals/jair/LaubliCNSST20,DBLP:journals/csur/MarufSH21}.

An obvious solution to this problem is to utilize context-aware MT models \cite{DBLP:conf/discomt/TiedemannS17}.
While document-level NMT models have been thoroughly studied in recent years, sentence-level MT remains the standard despite its inherent limitations.
One of the main reasons for this is that most of the document-level approaches rely on parallel training data with document-level metadata.
Most releases of large parallel training corpora lack this information and remain purely sentence-level \cite{DBLP:conf/acl/BanonCHHHEFKKKO20,DBLP:conf/acl/SchwenkWEGJF20}.
In contrast, large amounts of document-level monolingual data are readily available for almost all domains and languages.

In this work, we strive to build a context-aware MT system that does not rely on any parallel document-level training data.
Instead, we use monolingual documents to train a document-level language model (LM), which we fuse with an existing sentence-level MT model during translation.
While existing work on LM fusion shows that the fused model is able to incorporate document-level context \cite{DBLP:conf/acl-spnlp/JeanC20,DBLP:conf/naacl/SugiyamaY21}, these approaches can be improved.
Our work aims to do so in two main directions.

First, we acknowledge that NMT models implicitly learn the language modeling task during training.
Recently, \citet{christian-ilm} showed that estimating and neutralizing this internal LM can improve translation quality for sentence-level MT.
We adapt their approach to document-level LM fusion and demonstrate that this also improves discourse modeling.

Second, the contribution of the fused MT model, the document-level LM and the internal LM must be balanced by a set of fusion scales.
Existing work defines the fusion scales as static hyperparameters which are tuned on a validation set via an extensive grid search \cite{DBLP:journals/corr/GulcehreFXCBLBS15,DBLP:conf/acl-spnlp/JeanC20,DBLP:conf/naacl/SugiyamaY21}.
In our work, we provide two simple alternatives to grid search which allow for automatically tuned context-dependent fusion scales.
Our approaches eliminate the need for expensive tuning and further improve discourse-modelling.

The contributions of this work are as follows:
\begin{enumerate}
    \item We propose multiple extensions to the existing approaches on document-level LM fusion for MT.
    \item We compare our methods against two strong baselines: Back-translation, the to-date most popular way to utilize monolingual data for MT, and a task-specific LM re-ranking baseline for pronoun disambiguation.
    The comparison takes place over four diverse translation tasks in terms of general translation quality as well as specific context-dependant phenomena.
    \item We present first results on fusing a large language model (LLM) with a sentence-level MT system.
\end{enumerate}

\section{Related Works}

Most works on document-level NMT rely on parallel document-level data for system training.

\citet{DBLP:conf/discomt/TiedemannS17} propose to concatenate adjacent sentences on source and target side and input this into the NMT model which has the exact same architecture as the vanilla sentence-level transformer \cite{DBLP:conf/nips/VaswaniSPUJGKP17}.
Later, many works have proposed modifications to the architecture to better accommodate the additional context \cite{DBLP:journals/corr/JeanLFC17, DBLP:conf/naacl/BawdenSBH18, DBLP:conf/emnlp/ZhangLSZXZL18, DBLP:conf/acl/TitovSSV18, DBLP:conf/coling/KuangX18, DBLP:conf/emnlp/WerlenRPH18, DBLP:conf/acl/HaffariM18}.
However, it has been shown that the simple concatenation approach performs as good, if not better than these more complicated variants \cite{DBLP:conf/eamt/LopesFBZM20, DBLP:conf/acl/SunWZZHCL22}.

Maybe the biggest challenge for document-level NMT is that most of the parallel MT training data is not document-level \cite{DBLP:conf/mtsummit/Espla-GomisFRH19, DBLP:conf/acl/SchwenkWEGJF20}.
Recently there has been some effort to restore document-level meta information from existing sentence-level corpora but this is a very time consuming and error-prone process \cite{DBLP:conf/eacl/GhussinZG23}.
Therefore, approaches to document-level NMT have been proposed that utilize document-level monolingual data, of which typically large amounts are readily available.

One direction is to back-translate the document-level monolingual data to create synthetic parallel document-level data.
The reverse system used for back-translation can be either sentence-level \cite{DBLP:conf/wmt/Junczys-Dowmunt19, DBLP:conf/emnlp/SalehBCB19, DBLP:journals/corr/abs-2304-12959} or document-level \cite{DBLP:conf/discomt/SugiyamaY19, DBLP:conf/wmt/HuoHGDKN20}.
A downside of this approach is that the final MT system has to be re-trained to incorporate the new synthetic data.

Another line of work uses document-level language models in combination with sentence-level translation models.
\citet{DBLP:journals/corr/GulcehreFXCBLBS15} were the first to propose a log-linear combination of sentence-level language and NMT models, coining the term \lq{}shallow fusion\rq{}.
Recently, it was shown that the shallow fusion approach for sentence-level NMT can be improved by compensating for the implicitly learned internal language model of the NMT system \cite{christian-ilm}.
Regarding the integration of a document-level LM, earlier approaches simply use the LM for re-ranking the hypothesis of the sentence-level NMT model \cite{DBLP:conf/wmt/StahlbergSGB19, DBLP:journals/tacl/YuSSLKBD20}.
Several works have proposed to employ a log-linear combination between sentence-level NMT system and document-level LM \cite{DBLP:conf/discomt/GarciaCE19, DBLP:conf/acl-spnlp/JeanC20, DBLP:journals/corr/abs-2010-12827}.
Both \citet{DBLP:conf/acl-spnlp/JeanC20} and \citet{DBLP:journals/corr/abs-2010-12827} propose to also include the probabilities of the LM without context information in order to mitigate the influence of the current sentence on the LM probabilities.
While our approach also uses the output of a sentence-level LM, it is conceptually different from the previous works in that we want to mitigate the influence of the internal LM from the NMT model, resulting in a different final formulation.
To further improve LM incorporation, \citet{DBLP:conf/acl-spnlp/JeanC20} propose to use subword-dependent fusion scales instead of a single scale per model.

Apart from back-translation and LM integration there exist some other ways to utilize additional monolingual document-level data for MT.
\citet{DBLP:conf/emnlp/VoitaST19} train a document-level automatic post editing system on the monolingual data and use it to improve the hypotheses from a sentence-level NMT system in a two-pass approach.
Several works utilize the additional data in a multi-task learning approach \cite{DBLP:conf/wmt/Junczys-Dowmunt19} or for pre-training \cite{DBLP:conf/iclr/ZhuXWHQZLL20, DBLP:conf/ijcai/ChenLGDCLZZ21, DBLP:journals/tacl/LiuGGLEGLZ20, DBLP:conf/acl/ChenLGCLZZ20}.

Very recently, LLMs have shown their potential for the task of document-level NMT \cite{DBLP:journals/corr/abs-2304-02210}.
However, it is unclear how much parallel training samples were seen during the large scale pre-training on trillions of tokens.

\section{Document-level Language Model Fusion}

The sentence-level MT model translates a source sentence $ F $ into a target sentence $ E \coloneqq e_0^I $ of subwords $ e_i $.
In the document-level LM fusion approach, we additionally provide the $ k $ previous target-side sentences $ E_{-k}^{-1} $ as context\footnote{~At the beginning of the document we only provide as many sentences as available.}.

\subsection{Internal Language Model Neutralization}

As the translation model already implicitly learns probabilities that are source-independent, directly fusing the MT model and the document-level LM overvalues the source-agnostic probabilities.
Therefore, we estimate the internal LM of the MT model and in total combine three models during generation:
\begin{itemize}
    \item the existing sentence-level MT model $ p_{\text{TM}}(e_i) \coloneqq p_{\text{TM}}(e_i \smid e_{0}^{i-1}, F) $,
    \item the LM $ p_{\text{LM}}(e_i) \coloneqq p_{\text{LM}}(e_i \smid e_{0}^{i-1}, E_{-k}^{-1}) $ trained on monolingual documents with access to the previous target sentences $ E_{-k}^{-1} $,
    \item and a second LM $ p_{\text{ILM}}(e_i) \coloneqq p_{\text{ILM}}(e_i \smid e_{0}^{i-1}) $ which estimates the internal LM probabilities implicitly learned by the MT model.
    We train this LM separately on the target-side of the MT training data, as we found that this approach works best for document-level MT when compared to other approaches presented by \citet{christian-ilm}.
    This comparison can be found in Appendix~\ref{sec:internal-lm}.
\end{itemize}
We multiply the model output probabilities and normalize them.
The resulting probability distribution is now conditioned on both the source sentence  $ F$ and the target-side context $ E_{-k}^{-1} $:
\begin{align}
    \label{eq:log-linear-combination}
    p(e_i)
    &\coloneqq p(e_i \smid e_0^{i-1}, F, E_{-k}^{-1}) \nonumber \\
    &\coloneqq
    \frac{p_{\text{TM}}^{\lambda_0}(e_i) \cdot p_{\text{LM}}^{\lambda_1}(e_i) \cdot p_{\text{ILM}}^{-\lambda_2}(e_i)}{\sum_{e'} p_{\text{TM}}^{\lambda_0}(e') \cdot p_{\text{LM}}^{\lambda_1}(e') \cdot p_{\text{ILM}}^{-\lambda_2}(e')}.
\end{align}
Each model is weighted with a scalar $ \lambda_0, \lambda_1, \lambda_2 \geq 0 $, the internal LM is included with a negative exponent.
We tune these fusion scales on the validation set for \textsc{Bleu} via a grid search over $ \lambda_0, \lambda_1, \lambda_2 \in \{ 0, 0.1, \dots, 1 \} $.

Existing work on document-level LM fusion uses a similar formulation as our approach, but instead of neutralizing the internal LM of the MT model, it accounts for the sentence-level probabilities $ p_{\text{LM}}(e_i \smid e_0^{i-1}) $ of the document-level LM  \cite{DBLP:conf/acl-spnlp/JeanC20,DBLP:conf/naacl/SugiyamaY21}.
In the particular case where there are no previous sentences available, this approach simply falls back to using only the sentence-level MT model probabilities.
Our approach on the contrary can also leverage the gains obtained from sentence-level LM fusion and is theoretically more expressive.

\subsection{Context-dependent Fusion Scales}

Choosing appropriate fusion scales $ \lambda_0, \lambda_1, \lambda_2 $ in Equation~\ref{eq:log-linear-combination} is crucial.
Conventionally, the scales are tuned via grid search.
This is problematic in three aspects:
\begin{enumerate}
    \item Grid search is expensive.
    Testing e.g. ten possible values for each of the three model scales already requires translating the validation set 1000 times.
    \item The tuning process depends on the tuning data, its domain and the tuning objective.
    E.g., the scales that optimize document-targeted metrics differ from the ones that maximize sentence-level translation quality \cite{DBLP:conf/naacl/SugiyamaY21}.
    \item Fusion scales obtained by a hyperparameter grid search must be constant.
    Document-level context however is not uniformly useful for all predicted subwords.
\end{enumerate}
In the following, we propose two simple alternatives to obtaining fusion scales with grid search that overcome the aforementioned issues.

\subsubsection{On-the-fly Fusion Scales}

During decoding, the next subword $ e_i $ is chosen to maximize the fused probability (Equation~\ref{eq:log-linear-combination}).
We propose to also choose the fusion scales in a similar fashion and define them to maximize the fused model scores:
\begin{multline}
    (\lambda_0, \lambda_1, \lambda_2)
    \coloneqq 
    \\
    \argmax_{(\lambda_0, \lambda_1, \lambda_2)}
    \frac{p_{\text{TM}}^{\lambda_0}(e_i) \cdot p_{\text{LM}}^{\lambda_1}(e_i) \cdot p_{\text{ILM}}^{-\lambda_2}(e_i)}{\sum_{e'} p_{\text{TM}}^{\lambda_0}(e') \cdot p_{\text{LM}}^{\lambda_1}(e') \cdot p_{\text{ILM}}^{-\lambda_2}(e')}.
\end{multline}
Our model maximizes over the discrete set $ \lambda_0, \lambda_1, \lambda_2 \in \{ 0, 0.1, \dots, 1 \} $.
This approach obviates the need for separate scale tuning entirely and only has a small overhead during generation.

\subsubsection{Automatically Learned Fusion Scales}

Alternatively, we propose to learn the fusion scales automatically using a small amount of training examples $ (F, E, E_{-k}^{-1}) $ with document-level context, similarly to \citet{DBLP:conf/acl-spnlp/JeanC20}.
We obtain the training data by back-translating the monolingual data (see Section~\ref{sec:back-translation}).
Automatic learning allows us to implement subword-dependent fusion scales:
We introduce a set of learnable parameters $ \lambda_0(e), \lambda_1(e), \lambda_2(e) $ for each subword $ e $ from the target vocabulary and learn them automatically by optimizing the cross-entropy loss
\begin{multline}
    (\lambda_0, \lambda_1, \lambda_2)
    \coloneqq \argmax_{\lambda\colon V \to \mathbb{R}^3}
    \sum_{(F, E, E_{-k}^{-1})}
    \sum_i \\
    \log
    \frac{p_{\text{TM}}^{\lambda_0(e_i)}(e_i) \cdot p_{\text{LM}}^{\lambda_1(e_i)}(e_i) \cdot p_{\text{ILM}}^{-\lambda_2(e_i)}(e_i)}{\sum_{e'} p_{\text{TM}}^{\lambda_0(e')}(e') \cdot p_{\text{LM}}^{\lambda_1(e')}(e') \cdot p_{\text{ILM}}^{-\lambda_2(e')}(e')}.
    \label{eq:lm-learn-subword-dependent}
\end{multline}
Scale learning uses the same optimization parameters as the MT model was originally trained with.
The scale parameters are initialized with a small variance around zero while all other parameters are frozen.

\section{Document-level Language Model Pronoun Re-ranking}

Besides consistency, the main problem of discourse-modelling are ambiguities.
E.g. translating the English pronoun `it' to German requires access to the noun that it refers to, which might only be found in a preceding sentence \cite{DBLP:conf/nips/MullerKH19}.

We propose an approach specific to the En$\to$De language pair that directly targets the pronoun translation problem by re-ranking sentence-level hypotheses using a document-level LM.
We first translate each sentence independently using the sentence-level MT model. Each sentence-level translation is expanded to a set of candidates by replacing the pronouns with all alternatives (`er', `sie', `es').
All candidate translations are then scored in context of the preceding sentences using a document-level LM, and we select the pronoun for which the LM score is highest.

This approach is very much tailored to the specific pronoun translation problem for this specific language pair.
While it is theoretically possible to extend this approach to cover more cases, this will require extensive human effort and is probably not feasible in most scenarios.
However, we include it here, because it serves as a reasonable baseline for this popular pronoun translation benchmark.

\section{Document-level Back-translation}
\label{sec:back-translation}

The to-date most popular way of utilizing monolingual data for MT is to create synthetic parallel training data via back-translation \cite{DBLP:conf/acl/SennrichHB16}.
We train a sentence-level backwards MT system on the parallel data and use it to translate the document-level monolingual data back into the source language.
The sentence-level translations are concatenated to obtain synthetic parallel documents \cite{DBLP:conf/wmt/Junczys-Dowmunt19, DBLP:conf/emnlp/SalehBCB19, DBLP:conf/discomt/SugiyamaY19, DBLP:conf/wmt/HuoHGDKN20, DBLP:journals/corr/abs-2304-12959}.

To train the final systems we combine the authentic sentence-level parallel and the synthetic document-level data.
Combining both data sources is not straightforward, because of their varying size and the difference between sentence/document-level context.
Therefore, we first oversample the data accordingly to have roughly the same number of sentences in both parts.
Secondly, we turn the authentic sentence-level parallel data into `pseudo-documents' by concatenating them in a random order \cite{DBLP:conf/wmt/Junczys-Dowmunt19,DBLP:journals/corr/abs-1910-14075}.
This ensures that all training data has the same context size.
We found this procedure to perform best when incorporating synthetic document-level data.
For a detailed comparison, see Appendix~\ref{sec:back-translation-appendix}.

\section{Experiments}

\subsection{Tasks}

We evaluate our approaches on four different tasks of varying data conditions and domains.
Three tasks are on publicly available data and a fourth task is based on a large scale internal dataset in the e-Commerce domain.
All tasks include (sentence-level) parallel training data and document-level monolingual data from the same domain.
The exact data conditions are provided in Appendix~\ref{sec:training-data}.

The \emph{News En$\to$De} data consists of news articles while the \emph{TED En$\to$It} task consists of scientific talks.
Both are low resource with less than 1M training samples in total.
The \emph{Subtitles En$\to$De} data consists of subtitles from various TV shows and is medium size.
Finally, the \emph{e-Commerce En$\to$De} task is about translating item descriptions from e-Commerce listings and the training data is large scale with more than 100M examples.

While the parallel training data for the three academic tasks does provide document-level metadata, our approaches do not make use of this information and we assume that the parallel training data is sentence-level for most experiments.
We only make use of this information to provide a direct comparison against the setting where document-level parallel data is assumed to be available.
As ParaCrawl, like most other large-scale web-crawled parallel datasets, is not a document-level corpus, we can not conduct these experiments for the e-Commerce task.

We preprocess each corpus with byte-pair encodings \cite{DBLP:conf/acl/SennrichHB16} using the SentencePiece toolkit \cite{DBLP:conf/acl/Kudo18} learned on the parallel dataset with a shared vocabulary of 32k subwords (13.6k for TED).
For the e-Commerce task we additionally use inline casing \cite{DBLP:conf/wmt/BerardCR19,DBLP:conf/lrec/EtchegoyhenG20}.

\subsection{Settings}

We train transformer MT models in the `base' configuration \cite{DBLP:conf/nips/VaswaniSPUJGKP17}, implemented in Fairseq \cite{DBLP:journals/corr/abs-1904-01038}.
For the LMs we use a similar architecture but without the encoder.
Our document-level models use the same architecture as the sentence-level models, we simply include context sentences by concatenating the previous two source and target sentences to all training examples, separated by a reserved symbol \cite{DBLP:conf/discomt/TiedemannS17}.

Details on the optimization algorithm are given in Appendix~\ref{sec:training-data}.
The final model is selected based on the validation set perplexity.
We then perform beam search with beam size 12 and length normalization.
Document-level decoding uses the `last sentence' search strategy as described in \citet{christian-search}.

The document-level LMs are trained on a combination of target-side of the sentence-level parallel and document-level monolingual data.
Regardless of the task, we train the LMs for 300k update steps with batch size 90k, 10\,\% dropout, and 10\,\% label smoothing.

For the LM fusion experiments with non-static fusion scales, we restrict the search space to only consider scale combinations where $ \lambda_0 = 1 $ and $ \lambda_1 = \lambda_2 $.
A direction comparison is given in Appendix~\ref{sec:restricted}.
For back-translation, we use beam search with beam size 4 and increase the training time proportionally to the new data size.

\subsection{Evaluation}

Document-level evaluation is challenging, as intersentential context usually is only relevant for a small fraction of words.
Further, conventional metrics like \textsc{Bleu} \cite{DBLP:conf/acl/PapineniRWZ02} or \textsc{Comet} \cite{DBLP:conf/emnlp/ReiSFL20} do not appropriately measure how well document-level context is considered for those words where context does matter \cite{DBLP:conf/emnlp/LaubliS018,DBLP:journals/jair/LaubliCNSST20,DBLP:journals/csur/MarufSH21}.
However, we still report \textsc{Bleu} using Sacrebleu \cite{DBLP:conf/wmt/Post18} and \textsc{Comet}\footnote{~Using the \texttt{wmt22-comet-da} model \cite{DBLP:conf/emnlp/ReiSFL20}} on the task-specific in-domain test sets to evaluate the general MT quality.

To better evaluate the improvements from the document-level approaches, we focus on selected sentences for which document-level context is known to be important.
Here, we report on two test sets focusing on ambiguities.
The En$\to$De \emph{pronouns} test set released by \citet{DBLP:conf/wmt/MullerRVS18} was curated from OpenSubtitles shows and contains 12k examples.
Most examples require previous sentences as context to properly translate the English pronoun `it' with German `er', `sie' or `es'.
Further, the \emph{gender-referring professions} test sets released as contextual part of MT-GenEval \cite{DBLP:conf/emnlp/CurreyNPMLNHD22} are available for various target languages and focus on a wider range of ambiguous words, e.g. whether `the teacher' should be translated with `die Lehrerin' or `der Lehrer' in German.
Again, context from the previous sentences is required to determine the correct translation.
We use these test sets for En$\to$De and En$\to$It which both comprise approx. 1.1k examples that were created by translating Wikipedia articles.

Computing \textsc{Bleu} and \textsc{Comet} on these challenge test sets better reflects how well a MT system handles document-level context.
An even more specific metric can be obtained by focusing only on the ambiguous words.
Previous work commonly reports an accuracy metric that is based on contrastive scoring, which is computed by comparing the model probabilities of the reference against a set of contrastive examples \cite{DBLP:conf/wmt/MullerRVS18}.
This metric however can be misleading, as it not based on the generated translation but rather just on scoring.
MT systems with high contrastive scores often perform poorly when their generated hypothesis is evaluated \cite{DBLP:journals/corr/abs-2304-12959}.
Instead, we focus on translation-based document-targeted metrics.

On the pronouns test set, we compute a pronoun F1-score as proposed by \citet{christian-doc}.
This metric directly compares the pronouns of the hypothesis and the reference and is based on the \textsc{BlonDe} metric \cite{DBLP:conf/naacl/JiangLM0YHSCS022}.
On the professions test set, we report the translation-based accuracy metric suggested by their curators \cite{DBLP:conf/emnlp/CurreyNPMLNHD22}.
Further, for the Subtitles system we also report a formality F1-score on its test set as proposed by \citet{christian-doc}.

\subsection{Results}

We evaluate our approaches to utilize monolingual document-level data on the four MT tasks.
We apply them in two settings where a) we assume that all parallel data is purely sentence-level, \mbox{and b) also} the parallel data is document-level.

In an effort to compare to previous work, we re-implement LM fusion with static scales without subtracting the internal LM which was independently proposed by \citet{DBLP:conf/acl-spnlp/JeanC20} and \citet{DBLP:conf/naacl/SugiyamaY21}.
These works subtract the intersentential probabilities of the external LM instead.
Further, we also re-implement the non-static scales predicted with a `merging module' learned on parallel document-level data as proposed by \citet{DBLP:conf/acl-spnlp/JeanC20}.

We first evaluate our approaches on conventional metrics to measure their general MT performance.
Then, we focus on the document-targeted challenge sets to quantify how well they utilize document-level context.

\subsubsection{Conventional Metrics}
\begin{table*}
    \adjustbox{max width=\linewidth}{\begin{tabular}{|c|c||l||cc|cc|cc|cc|}
        \hline
        \multicolumn{2}{|c||}{Data} & \multirow{2}{*}{Method} & \multicolumn{2}{c|}{News} & \multicolumn{2}{c|}{Subtitles} & \multicolumn{2}{c|}{TED} & \multicolumn{2}{c|}{e-Commerce} \\
        \cline{1-2} \cline{4-11}
        parallel & mono. & & \BLEU & \COMET & \BLEU & \COMET & \BLEU & \COMET & \BLEU & \COMET \\
        \hline
        \multirow{12}{*}{sent.} & \multirow{2}{*}{-} & baseline (prev. work) & 32.8\footnote{\label{foot:baseline-a} External baseline by \citet{christian-search}} & - & 37.3\footnote{\label{foot:baseline-b} External baseline by \citet{DBLP:conf/wmt/HuoHGDKN20}} & - & 34.2\footnoteref{foot:baseline-a} & - & - & - \\
         & & baseline (ours) & 32.7 & 82.8 & 37.3 & 87.9 & 34.8 & 86.1 & 36.4 & 89.2 \\
        \cline{2-11}
        & \multirow{10}{*}{doc.} & {\scalebox{.7}[0.85]{(Jean, 2020; Sugiyama, 2021)}}\footnote{\label{foot:no-internal-lm} Re-implementation of LM fusion with neutralization of the intersentential LM probabilities instead of the internal LM, as introduced by \citet{DBLP:conf/acl-spnlp/JeanC20} and \citet{DBLP:conf/naacl/SugiyamaY21}} & 33.1 & 83.2 & 37.2 & 87.8 &  34.6 & 86.2 & 37.1 & 89.6 \\
        & & {\scalebox{.7}[0.85]{(Jean, 2020)}}\footnote{\label{foot:merging-module} Re-implementation of the `merging module' approach by \citet{DBLP:conf/acl-spnlp/JeanC20}. This approach uses parallel document-level data for scale learning.} & 32.9 & 83.0 & 37.3 & 87.9 & 34.5 & 86.2 & 36.6 & 89.2 \\
        \cline{3-11}
        & & LM: static & 34.8 & 84.2 & 37.2 & 87.8 & 34.9 & 86.2 & \textbf{37.3} & 89.6 \\
        & & LM: on-the-fly & 34.7 & 83.9 & 37.2 & 87.9 & 34.9 & 86.2 & 36.8 & \textbf{89.7} \\
        & & LM: auto. learned & 34.4 & 83.8 & 37.4 & 87.8 & 34.7 & 86.2 & 36.8 & 89.0 \\
        & & LM: re-rank pronouns & 32.6 & 82.7 & 36.9 & 87.8 & \multicolumn{2}{c|}{n.a.} & 36.4 & 89.2 \\
        \cline{3-11}
        & & back-translation & 37.1 & 85.2 & 37.2 & 87.6 & 35.1 & 86.6 & 36.2 & 89.3 \\
        & & ~+ LM: static & \textbf{37.4} & \textbf{85.6} & \textbf{37.6} & 87.7 & 35.2 & 86.6 & 35.0 & 88.9 \\
        & & ~+ LM: on-the-fly & 37.2 & 85.4 & 37.1 & 87.6 & 34.8 & 86.6 & 35.9 & 89.4 \\
        & & ~+ LM: auto. learned & 37.2 & 85.3 & 37.3 & 87.6 & 34.9 & 86.6 & 36.2 & 89.5 \\
        \hline
        \multirow{6}{*}{doc.} & - & baseline & 32.5 & 82.9 & 39.5 & \textbf{88.2} & \textbf{35.4} & 86.5 & \multicolumn{2}{c|}{\multirow{6}{*}{n.a.}} \\
        \cline{2-9}
        & \multirow{5}{*}{doc.} & LM: static & 35.1 & 84.3 & 38.9 & \textbf{88.2} &  35.2 & \textbf{86.7} & & \\
        & & LM: on-the-fly & 34.5 & 84.1 & 39.0 & 88.0 & 35.1 & \textbf{86.7} & & \\
        & & LM: auto. learned & 34.8 & 84.1 & 39.3 & 88.2 & 35.2 & 86.6 & & \\
        & & LM: re-rank pronouns & 32.3 & 82.8 & 39.1 & 88.1 & \multicolumn{2}{c|}{n.a.} & & \\
        & & back-translation & 37.2 & 85.3 & 37.5 & 87.8 & 34.6 & 86.5 & & \\
        \hline
    \end{tabular}}
    \centering
    \caption{
        Utilizing document-level monolingual data using different methods, reporting on the in-domain test sets of each task.
        \textsc{Bleu} and \textsc{Comet} are given in percentage. Best results for each column are highlighted.
    }
    \label{tab:comparison-bleu-comet}
\end{table*}

\begin{table*}
    \adjustbox{max width=\linewidth}{\begin{tabular}{|c|c||l||cc|ccc|c|cc|}
        \hline
        \multicolumn{2}{|c||}{Data} & \multirow{2}{*}{Method} & \multicolumn{2}{c|}{News} & \multicolumn{3}{c|}{Subtitles} & TED & \multicolumn{2}{c|}{e-Commerce} \\
        \cline{1-2} \cline{4-11}
        parallel & mono. & & pron. & proff. & pron. & proff. & form. & proff. & pron. & proff. \\
        \hline
        \multirow{12}{*}{sent.} & \multirow{2}{*}{-} & baseline (prev. work) & 45.3\footnoteref{foot:baseline-a} & - & 41.1\footnoteref{foot:baseline-a} & - & 59.4\footnoteref{foot:baseline-a} & - & - & - \\
        & & baseline (ours) & 45.1 & 65.9 & 41.7 & 65.3 & 57.2 & 65.4 & 42.6 & 63.7 \\
        \cline{2-11}
        & \multirow{11}{*}{doc.} & {\scalebox{.7}[0.85]{(Jean, 2020; Sugiyama, 2021)}}\footnoteref{foot:no-internal-lm} & 46.0 & 65.0 & 42.3 & 65.8 & 58.1 & 65.1 & 42.7 & 64.0 \\
        & & {\scalebox{.7}[0.85]{(Jean, 2020)}}\footnoteref{foot:merging-module} & 45.1 & 64.7 & 41.9 & 65.8 & 57.7 & 65.4 & 42.5 & 63.5 \\
        \cline{3-11}
        & & LM: static & 45.5 & 65.5 & 42.5 & 66.3 & 58.4 & 65.4 & 42.8 & 64.4 \\
        & & LM: on-the-fly & 48.0 & 65.5 & 44.2 & 65.9 & 58.9 & 66.4 & 44.4 & 66.2 \\
        & & LM: auto. learned & 46.7 & 64.9 & 42.8 & 65.5 & 58.6 & 65.6 & 44.0 & 65.2 \\
        & & LM: re-rank pronouns & 48.0 & 66.1 & 57.5 & 65.5 & 57.2 & n.a. & \textbf{54.5} & 64.0 \\
        \cline{3-11}
        & & back-translation & 48.7 & 80.5 & 52.3 & 67.0 & 58.5 & 65.1 & 42.9 & 67.1 \\
        & & ~+ LM: static & 48.5 & 80.6 & 53.1 & 68.3 & 53.8 & 65.4 & 42.6 & 66.0 \\
        & & ~+ LM: on-the-fly & 48.9 & \textbf{81.3} & 52.8 & 67.3 & 60.4 & 65.4 & 46.3 & \textbf{70.5} \\
        & & ~+ LM: auto. learned & 48.9 & 80.5 & 52.0 & 67.6 & 59.9 & 65.4 & 46.2 & 65.7 \\
        \cline{1-11}
        \multirow{6}{*}{doc.} & - & baseline & \textbf{55.9} & 71.2 & 67.2 & 70.8 & 61.9 & 67.2 & \multicolumn{2}{c|}{\multirow{6}{*}{n.a.}} \\
        \cline{2-9}
        & \multirow{5}{*}{doc.} & LM: static & 55.3 & 70.8 & 67.5 & 71.1 & 61.5 & 66.8 & & \\
        & & LM: on-the-fly & 55.8 & 72.3 & 67.8 & \textbf{71.9} & 61.4 & \textbf{67.6} & & \\
        & & LM: auto. learned & 55.7 & 71.5 & \textbf{67.4} & 71.0 & 61.6 & \textbf{67.6} & & \\
        & & LM: re-rank pronouns & 50.9 & 71.5 & 62.6 & 70.8 & 61.9 & n.a. & & \\
        & & back-translation & 52.1 & 79.4 & 62.8 & 67.3 & \textbf{62.0} & 65.7  & & \\
        \hline
    \end{tabular}}
    \centering
    \caption{
        Document-targeted evaluation of the different approaches utilizing document-level monolingual data.
        We report the pronoun F1 score \cite{christian-doc}, gender-referring professions accuracy \cite{DBLP:conf/emnlp/CurreyNPMLNHD22} and the formality F1 score on the Subtitles test set \cite{christian-doc}, all given in percentage. Best results for each column are highlighted.}
    \label{tab:comparison-doc-targeted}
\end{table*}

We start by evaluating on the in-domain test sets of the four MT tasks using the conventional MT metrics.
Here, we do not expect to see much improvements coming from the document-level context.
The results are presented in Table~\ref{tab:comparison-bleu-comet}.

Adding monolingual data gives the largest improvements on News and small improvements on the e-Commerce task.
On these two tasks, the monolingual data is in-domain and the improvements are likely because of the domain.
On Subtitles and TED we do not see any improvements as Subtitles already has a large amount of in-domain parallel data and the TED monolingual data is slightly out-of-domain.
We verified the domain effect by training sentence-level LMs on equal amounts of data from the target-side of the parallel and monolingual corpora and comparing their perplexities on the test sets.
Details are provided in Appendix~\ref{sec:domain-effects}.

None of the presented approaches significantly decreases translation performance in terms of conventional metrics.
The only exception is the back-translation which when added to the Subtitles and TED document-level baseline performs worse in \textsc{Bleu}.
In \textsc{Comet} however, this decrease is less prevalent.

\subsubsection{Document-targeted Metrics}

The results on the document-targeted test sets are shown in Table~\ref{tab:comparison-doc-targeted}.
First we discuss the scenario without access to document-level parallel training data.

\emph{LM fusion.}
Adding monolingual documents to the sentence-level baseline with the existing approaches from \citet{DBLP:conf/acl-spnlp/JeanC20} and \citet{DBLP:conf/naacl/SugiyamaY21} improves scores only marginally by on average +0.5\,\% absolute F1 score on the pronouns test set and no improvements on the professions set.
In comparison, our approach on LM fusion with the neutralization of the internal LM performs better:
E.g., the variant with on-the-fly scales on average improves the pronoun F1 score by +2.4\,\% and the professions accuracy by +0.9\,\%.
Compared to static scales, both on-the-fly and automatically learned scales yield small improvements and further do not involve the expensive grid search.

\emph{LM re-ranking pronouns.}
Our LM re-ranking approach was specifically tailored towards the pronouns test set.
We see most improvements on this test set, while the document-targeted metrics on the other test sets remain mostly unchanged.
For both the Subtitles and the e-Commerce task, LM re-ranking is the best approach of utilizing document-level monolingual data for this specific test set in the absence of document-level parallel data.
On News however, the gains are less prevalent:
Our analysis finds that even though the LM in this case can predict the pronouns correctly, the general translation quality of the baseline on this test set is low and therefore this model often fails to generate any pronouns at all.
This again highlights the discrepancy between scoring- and generation-based metrics.

\emph{Back-translation.}
In a direct comparison to LM fusion, back-translation outperforms LM fusion despite our improvements over the existing work.
Back-translation on average improves the pronouns F1 score by +4.8\,\% and the professions accuracy by +4.9\,\% over the sentence-level baseline.
This may also  highlight the importance of source-side document-level context as the LM based approaches do not have access to this.
Still, both back-translation and LM fusion can be combined and this yields further improvements:
The best performing approach not relying on document-level parallel data is to use both document-level back-translation and then LM fusion with on-the-fly scales, this method achieves on average +6.2\,\% F1 score on the pronouns and +6.0\,\% professions accuracy.

\emph{Parallel document-level data.}
The three baselines trained on parallel document-level data perform much better than the sentence-level baseline:
The document-level baselines score on average +18.0\,\% better on the pronouns F1 score and +4.2\,\% better on the professions accuracy than their sentence-level counterparts.
In addition, the systems trained on parallel documents also perform better than the sentence-level systems with additional monolingual documents in almost all cases.
This concludes that on these three tasks, having access to parallel document-level data is much more effective than utilizing monolingual document-level data, even though our monolingual corpora are much larger than the parallel ones.

Further including monolingual document-level data to the document-level baselines does not generally give additional improvements.
In particular, LM pronoun re-ranking decreases performance in this setting as the MT model itself is already better at predicting the correct pronoun than the LM trained on the document-level monolingual data.

\begin{table}
    \adjustbox{max width=\linewidth}{\begin{tabular}{|l||ccc|}
        \hline
        \multirow{2}{*}{Method} & \multicolumn{3}{c|}{contrastive pronoun acc.} \\
        \cline{2-4}
        & News & \!\!Subtitles\!\! & \!\!e-Comm.\!\! \\
        \hline
        sentence-level baseline & 49.0 & 46.4 & 46.1 \\
        \hline
        {\scalebox{.8}[0.9]{(Jean, 2020; Sugiyama, 2021)}}\footnoteref{foot:no-internal-lm} & 53.4 & 48.8 & 47.4 \\
        {\scalebox{.8}[0.9]{(Jean, 2020)}}\footnoteref{foot:merging-module} & 49.2 & 46.8 & 45.5 \\
        \hline
        LM: static & 55.2 & 49.5 & 48.5 \\
        LM: on-the-fly & 55.9 & 53.4 & 51.3 \\
        LM: auto. learned & 53.0 & 50.1 & 50.5 \\
        LM: re-rank pronouns & 65.7 & 73.9 & 64.8 \\
        \hline
        back-translation & 56.5 & 57.9 & 47.3 \\
        ~+ LM: static & 57.7 & 61.6 & 47.5 \\
        ~+ LM: on-the-fly & 57.9 & 61.1 & 54.3 \\
        ~+ LM: auto. learned & 56.7 & 59.0 & 54.1 \\
        \hline
        document-level baseline & \textbf{67.9} & \textbf{84.0} & n.a. \\
        \hline
    \end{tabular}}
    \centering
    \caption{Scoring-based, contrastive accuracies on the pronouns test set \cite{DBLP:conf/wmt/MullerRVS18} for the three En$\to$De tasks, reported in percent.}
    \label{tab:comparison-contrastive}
\end{table}

\emph{Contrastive scores.}
Previous work on document-level MT commonly evaluates document-level MT systems using contrastive scoring \citep[e.g.,][]{DBLP:conf/acl-spnlp/JeanC20, DBLP:conf/naacl/SugiyamaY21}.
As a direct comparison, we report the contrastive accuracies on the pronouns test set in Table~\ref{tab:comparison-contrastive}.
The trend is often similar to the translation-based metrics in Table~\ref{tab:comparison-doc-targeted}, however scoring-based improvements are much more pronounced.
Our experiments also show that strong contrastive accuracies do not necessarily lead to improvements on the generated hypothesis.
For example, on the News task, the contrastive scores of the LM pronoun re-ranking approach and the document-level baseline are similar but their translation-based scores differ strongly (c.f. Table~\ref{tab:comparison-doc-targeted}).

\subsubsection{Computational Cost}

\begin{table}
    \begin{tabular}{|l|c|c|}
        \hline
        \multirow{2}{*}{Method} & \multicolumn{2}{c|}{Time} \\
        \cline{2-3}
        & Preparation & Search \\
        \hline
        LM: static & 7187\,min & 5.4\,min \\
        LM: on-the-fly & \phantom{000}0\,min & 6.5\,min \\
            LM: auto. learned & \phantom{00}\nhphantom{.}8.3\,min & 5.4\,min \\
        \hline
    \end{tabular}
    \centering
    \caption{Total time necessary to tune different fusion scale variants on a single GPU, as well as the time spent during translation.
    We measure the time used to translate the News validation set.}
    \label{tab:compute}
\end{table}

We have shown that both the on-the-fly scales and the automatically learned scales improve document-targeted scores over static scores obtained via grid search.
Another downside of grid search is that the tuning process is quite expensive.
In Table~\ref{tab:compute}, we illustrate that a grid search with $ 11^3 $ parameters (as is used in this work) on a single GPU can easily take multiple days.
The on-the-fly scales do not require any preparation time as they are obtained entirely during search, in which the overhead is small.
The automatically scales on the other hand can be learned in just a few minutes and do not have any overhead in decoding.

\subsubsection{Large Language Model Integration}

\begin{table}
\adjustbox{max width=\linewidth}{\begin{tabular}{|l|c|c|c|c|}
    \hline
    \multirow{2}{*}{LM} & \multicolumn{2}{c|}{perplexity} & \multicolumn{2}{c|}{contrastive acc.} \\
    \cline{2-5}
     & news & \!e-comm.\! & pron. & proff. \\
    \hline
    NewsCrawl & 17.0 & 44.5 & 62.8 & \textbf{63.4} \\
    LLaMA & \textbf{9.2} & \textbf{11.8} & \textbf{80.0} & 62.3 \\
    \hline
\end{tabular}}
\centering
\caption{Comparing the small in-domain LM trained on NewsCrawl against the LLM LLaMA.}
\label{tab:llm-perplexity}
\end{table}

Recently, large language models (LLMs) which are trained on large corpora and long context sizes received a lot of attention \citep[e.g.,][]{DBLP:conf/nips/BrownMRSKDNSSAA20,DBLP:journals/corr/abs-2302-13971}.
In particular, they have also been able to perform document-level MT \cite{DBLP:journals/corr/abs-2301-07069,DBLP:journals/corr/abs-2302-09210,DBLP:journals/corr/abs-2304-03245,DBLP:journals/corr/abs-2304-02210}.
This raises the natural question whether LLMs can improve document-level LM fusion.

We experiment on the News task and compare our own small LM with 35M parameters trained on 2.2B tokens from the in-domain German NewsCrawl corpus against the 13B parameter version of LLaMA \cite{DBLP:journals/corr/abs-2302-13971}, which was trained on a total of 1000B tokens.
LLaMA's training data includes various domains and languages.
Only a small fraction of its data is German.
The small LM provides two sentences context while we query the LLM with 200 tokens context. 
We re-train our MT model and the small LM using the LLaMA tokenizer.
This leads to slightly worse performance compared to our previous experiments as the LLaMA tokenization was learned on general-domain English data.
For decoding we use a beam size of 4.

\begin{table}
\adjustbox{max width=\linewidth}{\begin{tabular}{|c|c||cc|cc|}
    \hline
    \multicolumn{2}{|c||}{LM Fusion} & \multicolumn{2}{c|}{news} & \multicolumn{2}{c|}{e-Commerce} \\
    \hline
    LM & Scales & \BLEU & \COMET & \BLEU & \COMET \\
    \hline
    (none) & - & 31.2 & 81.3 & 13.6 & 70.5 \\
    \hline
    \multirow{2}{*}{NewsCrawl} & static & 33.2 & 83.0 & 14.4 & 72.5 \\
     &  on-the-fly & 33.2 & 82.8 & 14.3 & 72.2 \\
     \hline
     \multirow{2}{*}{LLaMA} & static & \textbf{34.6} & \textbf{84.2} & \textbf{16.5} & \textbf{75.0} \\
     &  on-the-fly & 33.4 & 83.9 & 13.7 & 72.9 \\
    \hline
\end{tabular}}
\centering
\caption{Comparing fusion with a small LM and a LLM on general test sets.}
\label{tab:llm-conventional}
\end{table}

Table~\ref{tab:llm-perplexity} shows the perplexities of both LMs and their contrastive scores on the document-targeted test sets\footnote{~The professions test set was released without target-side context, which we therefore created ourselves by translating the source-side context with a commercial MT system.}.
Both LMs use the same vocabulary and thus their perplexities are comparable.
Because it is in general unclear whether test sets are or are not included in LLM training data, we also include the e-Commerce test set which was translated by ourselves for the purpose of cross-validation.
On both test sets, the LLM perplexities are much better than the ones of the small in-domain LM.
LLaMA's contrastive scores are also much better on the pronouns test set.

\begin{table*}
    \adjustbox{max width=\linewidth}{\begin{tabular}{|c|c|c||c|cc|cc|}
    \hline
    \multicolumn{3}{|c||}{Fusion Scales Learning} & \multirow{2}{*}{$ \lambda $} & \multicolumn{2}{c|}{valid set} & \multicolumn{2}{c|}{doc.-targeted} \\
    \cline{1-3} \cline{5-8}
    Scales & Crit. & Train Set & & \BLEU & \COMET & pron. & proff. \\
    \hline
    none & - & - & 0.0 & 24.5 & 80.9 & 45.1 & 65.9 \\
    \hline
    \multirow{3}{*}{\shortstack{subword- \\ agnostic}} & grid search & valid set & 0.40 & 25.4 & 81.8 & 46.5 & 65.1 \\
    \cline{2-8}
    & \multirow{2}{*}{CE} & valid set & 0.34 & 25.5 & 81.7 & 46.4 & 65.0 \\
    & & synthetic & 0.46 & 25.3 & 81.6 & \textbf{47.1} & 65.2 \\
    \hline
    \multirow{2}{*}{\shortstack{subword- \\ dependent}} & \multirow{2}{*}{CE} & valid set & - & \textbf{26.6} & \textbf{81.8} & 46.1 & 65.0 \\
      & & synthetic & - & 25.4 & 81.6 & 46.9 & 65.4 \\
      \hline 
    \end{tabular}}
    \centering
    \caption{Automatically learning subword-dependent and -agnostic fusion scales on the News task. We employ the restriction $ \lambda_0 \coloneqq 1$ , $ \lambda \coloneqq \lambda_1 = \lambda_2 $.}
    \label{tab:learned-weights-data}
\end{table*}

\begin{table}
\adjustbox{max width=\linewidth}{\begin{tabular}{|c|c||ccc|}
  \hline
  \multicolumn{2}{|c||}{LM Fusion} & \multicolumn{3}{c|}{document-targeted} \\
  \hline
  LM & Scales & pron. & proff. & form. \\
  \hline
  (none) & - & 44.5 & 65.7 & 33.4 \\ 
  \hline
  \multirow{2}{*}{NewsCrawl} & static & 46.3 & 66.3 & 34.7 \\
   & on-the-fly & 47.4 & 66.7 & 34.3 \\
   \hline
   \multirow{2}{*}{LLaMA} & static & \textbf{51.6} & 66.9 & \textbf{36.1} \\
   & on-the-fly & 48.2 & \textbf{68.9} & 35.2 \\
  \hline
\end{tabular}}
\centering
\caption{Fusion with a small LM against a LLM, reporting the translation-based scores on the document-targeted test sets.}
\label{tab:llm-doc-targeted}
\end{table}

Table~\ref{tab:llm-conventional} shows the performance of LM fusion with the two LMs in \textsc{Bleu} and \textsc{Comet}.
Both LMs notably improve translation, but the LLM translation quality is best.
Fusion with LLaMA yields +3.4\,\% absolute improvements on the in-domain test set.
Improvements on the e-Commerce test set are similar, indicating that the gains are not an effect of data leakage of the test set into the training data.
While the on-the-fly scales and the static scales perform similarly for the small LM, on-the-fly scales do not perform as well for the LLM.

The improvements measured on the in-domain test sets are likely not because of document-level context but rather due to the increased amount of data.
Therefore, we continue our evaluation with the document-targeted scores.
Table \ref{tab:llm-doc-targeted} depicts the results.
On these metrics, the LLM outperforms the small LM by an even larger margin.
In general, the improvements are correlated to their contrastive scores (c.f. Table~\ref{tab:llm-perplexity}).

\subsection{Extended Analysis on Automatically Learned Fusion Scales}

In our experiments we use the validation set of the News task to find the best working methods.
We share some insights in the following. 

\begin{figure}
    \centering
    \adjustbox{max width=\linewidth}{\input{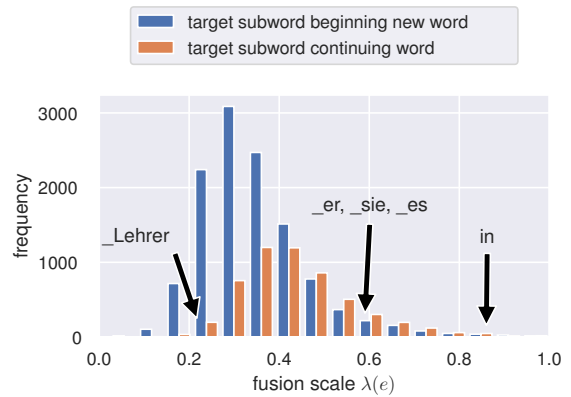}}
    \vspace{0.1cm}
    \caption{Distribution of the automatically learned LM fusion scales for different target-side subwords on the News task.
    Subwords for which document-level context is often necessary, such as the German pronouns `\char`_er', `\char`_sie', `\char`_es', and the suffix `in' marking female professions, have learned higher scales than nouns like `\char`_Lehrer'.}
    \label{fig:learned-weights}
\end{figure}

\emph{How are the automatically learned scales distributed?}
Figure~\ref{fig:learned-weights} shows the distribution of the automatically learned scales for the News task.
The learned LM scale of subwords that continue another subword are in general higher than the ones that begin a new word.
This is intuitive as continuing a subword is an LM task while beginning a new word requires information about the source sentence.

\emph{How much data is needed for automatically learning scales?}
The static fusion scales are usually tuned on a small validation set via grid search.
Table~\ref{tab:learned-weights-data} shows that it is also possible to use automatic differentiation to learn static scales only on the validation set.
The automatically learned subword-agnostic scales have similar values as the ones tuned via grid search and therefore also their translation performance is similar.
Learning subword-dependent scales automatically on the validation set on the other hand improves performance on this set, but does not generalize which indicates overfitting.

\section{Conclusions}

This work presents multiple extensions to document-level LM fusion, a technique of utilizing document-level monolingual data for context-aware MT.
In comparison to existing work, our extensions significantly improve discourse-modeling across four MT tasks and furthermore are computationally more efficient.
We conduct evaluations against two baselines: document-level back-translation and a task-specific LM re-ranking method.
Despite our extensions, back-translation in general still outperforms document-level LM fusion.
Nevertheless back-translation can be effectively combined with LM fusion, further improving translation performance.
On very specific test sets, the LM re-ranking performs best.
However, our experiments also show that systems trained on document-level parallel data outperform the best systems trained with monolingual documents only.

Finally, this work is the first to explore document-level LM fusion with LLMs.
First findings demonstrate that fusion with an LLM outperforms a small LM trained on in-domain data and open the path for future investigations.

\section*{Limitations}

The experiments in this work were limited to four MT tasks, from which two are low-resource and three are translating from English into German.
Apart from the experiments with the LLM, we did not conduct any experiments on a large-scale dataset of multi-domain monolingual documents.
The LLM in our experiments only has 7B parameters, while much larger LLMs exist \cite[e.g.,][]{DBLP:journals/corr/abs-2302-13971}.

Further, our work focuses only on one specific architecture for document-level MT and uses only two sentences target-side context.
Various other architectures exist and may entail different properties.
This work further does not investigate the behavior of larger translation models.

Another limitation lies in the evaluation of document-level MT models.
The document-level targeted metrics we used are all reference-based and limited to the translation of pronouns, gender-referring professions or salutation forms.
Other discourse phenomena like e.g. cohesion exist \cite{DBLP:journals/csur/MarufSH21} but were not studied in our work.
It is unclear how well automated metrics actually correlate with the actual document-level translation quality \cite{DBLP:conf/emnlp/CurreyNPMLNHD22}, and this work did not perform any qualitative analysis.



\bibliography{anthology,custom}
\bibliographystyle{acl_natbib}

\newpage

\newpage
\appendix

\section{Appendix}

\begin{table*}
\adjustbox{width=\textwidth}{\begin{tabular}{|l||ccc|ccc|cc|ccc|}
    \hline
    \multirow{2}{*}{\shortstack[l]{Data}} & \multicolumn{3}{c|}{News} & \multicolumn{3}{c|}{Subtitles} & \multicolumn{2}{c|}{TED} & \multicolumn{3}{c|}{e-Commerce} \\
     & test & pron. & proff. & test & pron. & proff. & test & proff. & test & pron. & proff. \\
    \hline
    parallel data & 129.7 & 125.7 & 168.6 & \textbf{26.6} & \textbf{36.0} & \textbf{103.2} & \textbf{47.6} & \textbf{114.8} & 61.7 & \textbf{44.2} & \textbf{52.6} \\
    monolingual data & \textbf{97.1} & \textbf{94.9} & \textbf{161.3} & 27.8 & 36.6 & 117.4 & 75.4 & 116.1 & \textbf{50.8} & 48.7 & 57.5 \\
    \hline
\end{tabular}}
\centering
\caption{Perplexities of sentence-level LMs trained on equal amount of target-side data.}
\label{tab:setup-domain-effects}
\end{table*}

\subsection{Model Training}
\label{sec:training-data}

\emph{Data.}
The \emph{News En$\to$De} task comprises 330k parallel sentences from NewsCommentary v14\footnote{~\url{https://data.statmt.org/news-commentary/v14/}}, which we combine with document-level monolingual data from NewsCrawl\footnote{~\url{https://data.statmt.org/news-crawl/}} (70M sentences\footnote{~To reduce training time, our back-translation experiments on this task utilize only the first 2M sentences.}).
Our \emph{Subtitles En$\to$De} data consists of a total of 39M monolingual movie show subtitles from OpenSubtitles, from which a subset of 22.5M sentences has been aligned to English sentences and forms our parallel training data \cite{DBLP:conf/lrec/LisonTK18}.
For \emph{TED En$\to$It} we use 230k parallel sentences from scientific TED talks released as part of the IWSLT17 multilingual task \cite{DBLP:conf/iwslt/CettoloFBNSSYF17} which we combine with 2.2M sentences of talks from the European parliament \cite{DBLP:conf/mtsummit/Koehn05}.
Finally, the \emph{e-Commerce En$\to$De} task is about translating item descriptions from e-Commerce listings.
We use 326M parallel sentences of out-of-domain parallel training data from the ParaCrawl v9 corpus \cite{DBLP:conf/mtsummit/Espla-GomisFRH19} which we combine with 128k parallel sentences in-domain data.
The monolingual data was sampled from item descriptions and is entirely in-domain (119M sentences).

The sizes of our training corpora are shown in Table~\ref{tab:dataset}.

\begin{table}
    \adjustbox{max width=\linewidth}{\begin{tabular}{|cc||ccc|}
        \hline
        Task & Data & docs & sents & words \\
        \hline
        \multirow{2}{*}{News} & parallel & 8.5k & 330k & 7.4M \\
        & mono. & 3M & 70M & 1.0B \\
        \hline
        \multirow{2}{*}{Subtitles} & parallel & 30k & 22.5M & 136M \\
        & mono. & 47k & 39M & 223M \\
        \hline
        \multirow{2}{*}{TED} & parallel & 1.9k & 230k & 3.7M \\
        & mono.  & 6k & 2.2M & 54.6M \\
        \hline
        \multirow{2}{*}{e-Commerce} & parallel & n.a. & 326M & 9.6B \\
        & mono. & 1.5M & 119M & 3.1B \\
        \hline
    \end{tabular}}
    \centering
    \caption{Training data statistics.}
    \label{tab:dataset}
\end{table}

On each task, we use a validation set for selecting the best checkpoint, tuning the fusion scales and for finding which method works best.
For the final comparison in Table~\ref{tab:comparison-bleu-comet} we then report on an unseen test set of the same domain.

The News validation set is \texttt{newstest2015}, and \texttt{newstest2018} as test set.
For Subtitles, our validation and test sets were sampled from the training corpus.
The precise document IDs for the validation set are:
1995/254, 1997/165, 2000/313, 2002/461, 2005/441, 2007/781, 2010/273, 2012/757, 2015/1488, 2017/525 for the validation set; and for our test set: 1997/310, 2002/40, 2007/189, 2012/1085, 2017/644.
The test set is the same as used in \citet{DBLP:conf/wmt/HuoHGDKN20}.
For TED, we concatenate \texttt{dev2010} and \texttt{tst2010} and use \texttt{tst2017.mltlng} as test set.
For e-Commerce, we create the validation and test set ourselves by translating English e-Commerce item descriptions into German:
Our validation set comprises 85 documents (2882 sentences) and the test set 100 documents (2520 sentences).

As the pronouns test set \cite{DBLP:conf/wmt/MullerRVS18} was extracted from the OpenSubtitles corpus, we remove these sentences from the Subtitles training data.
The professions test set \cite{DBLP:conf/emnlp/CurreyNPMLNHD22} was curated from Wikipedia articles and is not part of our training corpora.

\emph{Models.}
We train the News, Subtitles and TED models with a shared embedding and projection matrix.
Th resulting MT models for News and Subtitles have 60M parameters, 51M parameters for TED and 90M for e-Commerce.
For model training we use eight Tesla V100-SXM2-32GB GPUs.
Training the baselines takes approximately 7h for News, 21h for Subtitles, 5h for TED, and 30h for e-commerce.
Due to resource constraints, we report only a single run for each experiment.

\emph{Optimization.}
For optimization we use Adam \cite{DBLP:journals/corr/KingmaB14} and a batch size of 22k subwords.
The low-resource MT models (News, TED) are trained for 100k update steps with 30\,\% dropout, 20\,\% label smoothing and weight decay, while the high-resource models (Subtitles, e-Commerce) are trained for 300k updates with 10\,\% dropout, 10\,\% label smoothing and no weight decay.

\subsection{Domain Effects}
\label{sec:domain-effects}

In an effort to estimate how well the domain of the training data matches the test sets, we train LMs on the target-side part of the parallel and the monolingual training data.
Within each task, the LMs are trained with the same parameters and the same vocabulary.
We then report the perplexities on the task-specific test sets and the document-targeted challenge sets in Table~\ref{tab:setup-domain-effects}.

For News, the monolingual data is more in-domain for all test sets.
Similarly the domain of the e-Commerce monolingual data is closer to the task-specific test set.
For Subtitles, the domains of parallel and monolingual data are more or less equal and on TED, the monolingual data is slightly out-of-domain.

This domain effect explains the improvements in \textsc{Bleu} and \textsc{Comet} on the task-specific test sets that we reported in Table~\ref{tab:comparison-bleu-comet} on News and on e-Commerce.

\subsection{Comparing Internal Language Model Estimations}
\label{sec:internal-lm}

\begin{table}
    \adjustbox{max width=\linewidth}{\begin{tabular}{|l||cc|cc|}
        \hline
        \multirow{2}{*}{Approach} & \multicolumn{2}{c|}{\multirow{1}{*}{valid set}} & \multicolumn{2}{c|}{\multirow{1}{*}{doc.-targeted}}  \\
        \cline{2-5}
        & \BLEU & \COMET & pron. & proff. \\
        \hline
        baseline & 24.5 & 80.9 & 45.1 & \textbf{65.9} \\
        \hline
        LM fusion & 24.8 & 81.2 & 45.0 & 65.8 \\
        \,+~{\scalebox{.7}[0.85]{(Jean, 2020; Sugiyama, 2021)}}\footnoteref{foot:no-internal-lm} & 24.9 & 81.2 & 46.0 & 65.0 \\
        \,+~ILM: separate & \textbf{25.8} & \textbf{82.1} & \textbf{47.3} & 65.2 \\
        \,+~ILM: $ h = 0 $ & 25.5 & 82.0 & 43.8 & 64.7 \\
        \,+~ILM: mini self-att. & \textbf{25.8} & \textbf{82.1} & 44.6 & 65.1 \\
        \hline
    \end{tabular}}
    \centering
    \caption{Document-level LM fusion (a) without subtracting any LM, (b) subtracting the sentence-level probabilities of the external LM \cite{DBLP:conf/acl-spnlp/JeanC20,DBLP:conf/naacl/SugiyamaY21}, and (c) subtracting different approximations of the internal LM (ILM) learned by the MT model, reported on the News task.}
    \label{tab:ilm-subtraction}
\end{table}

\citet{christian-ilm} propose several ways of approximating the internal LM learned implicitly by the MT model in the context of sentence-level MT.
We evaluate three of their approaches for document-level LM fusion and compare them against the existing document-level LM fusion approach that subtracts the sentence-level probabilities of the external LM \cite{DBLP:conf/acl-spnlp/JeanC20,DBLP:conf/naacl/SugiyamaY21}.
Table~\ref{tab:ilm-subtraction} shows the results:
Subtracting the internal LM substantially improves LM fusion over existing work.
Estimating it by training a separate LM on the same data as the MT model works best.

\subsection{Fusion Scale Restrictions}
\label{sec:restricted}

\begin{table}
\adjustbox{max width=\linewidth}{\begin{tabular}{|c|c||cc|cc|}
  \hline
  \multicolumn{2}{|c||}{Fusion Scales} & \multicolumn{2}{c|}{valid set} & \multicolumn{2}{c|}{doc.-targeted}  \\
  \hline
  Approach & Restriction & \BLEU & \COMET & pron. & proff. \\
  \hline
  none & - & 24.5 & 80.9 & 45.1 & 65.9 \\
  \hline
  \multirow{2}{*}{static}  & - & \textbf{25.8} & \textbf{82.1} & 47.3 & 65.2 \\
   & $ \lambda_0 \!=\! 1, \lambda_1 \!=\! \lambda_2 $ & 25.4 & 81.8 & 46.5 & 65.1 \\
  \hline
  \multirow{2}{*}{on-the-fly} & - & 22.3 & 78.3 & 43.4 & 69.3 \\
  & $ \lambda_0 \!=\! 1, \lambda_1 \!=\! \lambda_2 $ & 25.6 & 81.8 & \textbf{48.0} & 65.5 \\
  \hline
  auto. & - & 24.8 & 80.7 & 44.7 & \textbf{69.4} \\
  learned & $ \lambda_0 \!=\! 1, \lambda_1 \!=\! \lambda_2 $ & 25.3 & 81.5 & 46.7 & 64.9 \\
  \hline
\end{tabular}}
\centering
\caption{LM fusion with an imposed restriction on the search space of the fusion scales $ \lambda_0, \lambda_1, \lambda_2 $, reported on the News task.}
\label{tab:scale-restrictions}
\end{table}

The three LM fusion scales $ \lambda_0, \lambda_1, \lambda_2 $ in Equation~\ref{eq:log-linear-combination} balance the contribution of the MT model and the two LMs.
In our experiments the optimal scales usually lie at $ \lambda_0 \approx 1 $ and $ \lambda_1 \approx \lambda_2 $.
This is plausible as the internal LM ($ \lambda_2 $) should neutralize the external LM ($ \lambda_1 $) to the same degree.
For the non-static fusion scales however, we find that searching over the three-dimensional search space of independent $ \lambda_0, \lambda_1, \lambda_2 $ finds unintuitive scale combinations and that this causes bad performance.
Therefore in our experiments we restrict the search space of fusion scales to the one-dimensional slice where $ \lambda_0 = 1 $ and $ \lambda_1 = \lambda_2 $.
Table~\ref{tab:scale-restrictions} gives a direct comparison.

\subsection{Document-level Back-translation}
\label{sec:back-translation-appendix}

In Table~\ref{tab:bt-doc-level-hyperparameter} we compare back-translation using document-level data against sentence-level back-translation.

\begin{table}
  \adjustbox{max width=\linewidth}{\begin{tabular}{|c|c||cc|cc|}
      \hline
      \multicolumn{2}{|c||}{Data} & \multicolumn{2}{c|}{valid set} & \multicolumn{2}{c|}{doc.-targeted} \\
      \hline
      parallel & mono. & \BLEU & \COMET & pron. & proff. \\
      \hline
      sent. & - & 24.5 & 80.9 & 45.1 & 65.9 \\
      sent. & sent. & 27.0 & \textbf{83.2} & 46.7 & 65.7 \\
      \hline
      sent. & doc. & 26.9 & 82.4 & 47.8 & \textbf{80.7} \\
      pseudo-doc. & doc. & \textbf{27.1} & 83.0 & \textbf{48.7} & 80.5 \\
      \hline
  \end{tabular}}
  \centering
  \caption{Effect of back-translation on the News task.}
  \label{tab:bt-doc-level-hyperparameter}
\end{table}

Sentence- and document-level back-translation gives the same performance improvements in \textsc{Bleu} and \textsc{Comet}, however only back-translation on document-level improves the document-targeted metrics.
For document-level back-translation we find that creating pseudo-documents from the parallel data is necessary to achieve the same \textsc{Bleu} and \textsc{Comet} scores as sentence-level back-translation.

\end{document}